\documentclass[preprint,12pt]{elsarticle}
\usepackage{amssymb}
\usepackage{amsmath}
\usepackage[utf8]{inputenc}
\usepackage{hyperref}
\usepackage{tabularx}
\usepackage{booktabs}
\usepackage{longtable}

\journal{Fuel}

\begin{document}

\begin{frontmatter}

\title{A Generative Deep Learning Workflow for Inverse Molecular Design of Fuels}

\author[label1]{Kiran K. Yalamanchi}
\author[label1]{Pinaki Pal\corref{cor1}}
\ead{pal@anl.gov}
\cortext[cor1]{Corresponding author}
\author[label2]{Balaji Mohan}
\author[label3]{Abdullah S. AlRamadan}
\author[label2]{Jihad A. Badra}
\author[label4]{Yuanjiang Pei}

\affiliation[label1]{
   organization={Transportation and Power Systems Division, Argonne National Laboratory}, 
   addressline={9700 S Cass Avenue}, 
   city={Lemont}, 
   postcode={IL 60439}, 
   country={USA}
}

\affiliation[label2]{
   organization={Digital R\&D Division, Research and Development Center, Saudi Aramco}, 
   city={Dhahran}, 
   country={Saudi Arabia}
}

\affiliation[label3]{
   organization={Transport Technologies Division, Research and Development Center, Saudi Aramco}, 
   city={Dhahran}, 
   country={Saudi Arabia}
}

\affiliation[label4]{
   organization={Aramco Americas: Aramco Research Center–Detroit}, 
   addressline={Novi, MI}, 
   country={USA}
}

\begin{abstract}
In the present work, a generative deep learning framework combining a Co-optimized Variational Autoencoder (Co-VAE) with quantitative structure-property relationship (QSPR) techniques is developed to enable inverse molecular design of fuels. The Co-VAE approach integrates an auxiliary fuel property prediction regression head with the VAE latent space, enhancing molecular reconstruction and accurate property estimation (Research Octane Number (RON) chosen as the fuel property of interest for demonstration studies). A subset of the GDB-13 database, combined with a curated RON database, is used for the Co-VAE training. Hyperparameter tuning is further utilized to optimize the balance among reconstruction fidelity, chemical validity, and RON prediction. Subsequently, an independent regression model is trained to further improve RON prediction accuracy, and a differential evolution algorithm is employed to efficiently navigate the Co-VAE latent space and identify promising fuel molecule candidates with RON greater than a chosen threshold. The overall generative deep learning framework captures complex structure-property relationships within a latent representation, and can be readily extended to different or multiple fuel properties, allowing exploration of large chemical spaces relevant to fuel design. Furthermore, the framework can be further augmented by incorporating additional synthesizability criteria to improve applicability and reliability for de novo design of novel high-performance fuels.
\end{abstract}


\end{frontmatter}

\section{Introduction}
\label{sec:intro}
Advancements in automotive technology and stricter environmental and efficiency standards necessitate novel fuel formulations that enhance engine efficiency while minimizing emissions~\cite{Ref33}. Modern engines require fuels with high knock resistance and cleaner combustion characteristics, but the chemical space for viable fuel molecules is vast. Traditional methods of fuel design and development rely on experimental trial-and-error and expert intuition. Although such methods remain essential for concrete validation, they are inefficient for exploring large chemical spaces of potential fuel molecules~\cite{Ref34}.

Fuel design has long been supported by model-based approaches. Systematic methodologies for designing tailor-made fuel components and blended products, for example, group-contribution methods and blend-design algorithms to match target specifications, have been applied to gasoline fuels and their blends~\cite{Ref31}. These approaches are effective for constrained formulation problems but do not, by themselves, propose novel molecular structures across large chemical spaces. In this context, machine learning (ML)--aided molecular design in domains such as drug and material discovery has matured~\cite{Ref32}, and similar concepts have increasingly been explored for fuel applications. ML-based Quantitative Structure-Property Relationship (QSPR) models have been developed for predicting fuel properties, where methods based on molecular fingerprints and graph representations have improved octane prediction performance~\cite{Ref9,Ref10,Ref11}. However, while advanced ML methods can enhance prediction accuracy over existing model-based approaches, many QSPR models are trained on relatively limited datasets with hand-crafted features, which may not generalize across broad chemical spaces.

Generative deep learning offers a complementary route to inverse fuel design by learning continuous representations from which novel structures can be sampled~\cite{Ref32}. In drug discovery, Variational Autoencoders (VAEs), Generative Adversarial Networks (GANs), and reinforcement learning (RL) frameworks have been employed to generate property-conditioned candidates~\cite{Ref1,Ref2,Ref3,Ref4}. Similar ideas have been transferred to fuel design: Liu \emph{et al.}~\cite{Ref5} used multi-objective imitation learning for low-data fuel design; Rittig \emph{et al.}~\cite{Ref6} combined graph generative models with optimization for high-octane species; Kuzhagaliyeva \emph{et al.}~\cite{Ref7} extended artificial intelligence (AI)-driven design to fuel mixtures; and Yalamanchi \emph{et al.}~\cite{Ref8} incorporated uncertainty quantification in related fuel property prediction settings. Liu \emph{et al.}~\cite{Ref12} further developed a VAE that jointly optimizes reconstruction and an embedded property predictor for fuel-relevant targets. These studies show that generative models can explore the chemical space more directly than QSPR-only workflows, but integrated demonstrations on fuel-relevant corpora with explicit inverse design for desirable fuel properties remain limited.

In authors’ previous work~\cite{Ref13}, a Long Short-Term Memory (LSTM)-based VAE framework was developed that learns a continuous latent representation of molecular structures suitable for generation and prediction. The VAE was trained on a large chemical database using one-hot-encoded Simplified Molecular Input Line Entry System (SMILES) strings as input. Strategies to balance reconstruction fidelity and latent space regularization through weightage of the regularization term ($\beta$) were systematically investigated. It was found that gradually increasing $\beta$ to a moderate value (0.25) provided the best trade-off between reconstruction accuracy and latent space consistency. Higher $\beta$ values risked posterior collapse—where the decoder ignores latent features—while lower $\beta$ values led to poor regularization. Advanced techniques, such as cyclical $\beta$ schedules~\cite{Ref14}, hierarchical priors~\cite{Ref15}, and penalization of latent variable correlations~\cite{Ref16}, were also explored. The results indicated that although a hierarchical latent prior offered some improvements in reconstruction accuracy, its benefits were modest, and substantial penalization of latent total correlation degraded performance. Overall, employing a standard VAE with an annealed $\beta$ schedule captured most of the improvements.

Demonstrations of integrated data-driven fuel design workflows that connect fuel property-conditioned generative AI modeling with effective latent-space search in fuel-relevant molecular structural  space remain limited. To this end, this work presents a streamlined VAE-based pipeline for inverse fuel design along with a use case demonstration for target fuel property of Research Octane Number (RON). In particular, a co-optimized VAE (Co-VAE) approach is first employed that directly incorporates fuel property information into the generative model through an auxiliary fuel property regression head trained on the labeled subset to learn meaningful latent molecular structural representations for fuel property prediction. Systematic hyperparameter optimization (HPO) of the Co-VAE architecture and training parameters is performed using Bayesian optimization to improve reconstruction quality and property-prediction accuracy. Thereafter, a QSPR modeling step is integrated by mapping the learned latent representations to the target fuel property (Research Octane Number (RON) in this case), constructing a separate fuel property regression model for fine-tuning. Multiple ML algorithms are evaluated for this task, applying systematic hyperparameter searches to identify the optimal fuel property predictor. Finally, latent-space screening based on differential evolution, together with chemical validity checks, is used to identify promising fuel molecules with the desired target property.

\begin{figure}[ht]
    \centering
    \includegraphics[width=1\textwidth]{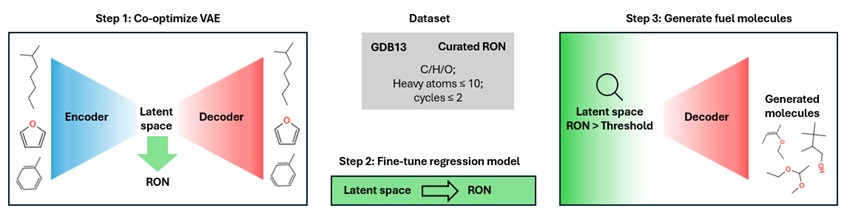} 
    \caption{Integrated workflow for generative AI-driven fuel design with RON as the fuel property of interest. Step 1: Co-optimize the VAE for molecular reconstruction and RON prediction. Step 2: Fine-tune the regression model to enhance RON prediction. Step 3: Explore the VAE latent space regions linked to high RON and decode the corresponding species.}
    \label{fig1}
\end{figure}

This integrated approach, shown in Figure~\ref{fig1}, leverages the strengths of both generative and regression models, facilitating efficient exploration of the chemical space to design novel fuel molecules with desired properties. By combining a VAE-based generative model with a data-driven property predictor and employing systematic optimization of the pipeline components, the present work aims to demonstrate that such an integrated workflow can be applied to fuel-relevant data and yield high--performance fuel candidates suitable for downstream evaluation. 

\section{Co-VAE Model Development}
\label{sec:covademodel}
\subsection{Data Curation and Processing}
The datasets utilized in this study are fundamental to developing the Co-VAE model and predicting RON. Building upon the authors' previous work~\cite{Ref13}, a refined subset of the GDB-13 database~\cite{Ref17} was employed, which is an extensive collection of small organic molecules containing up to 13 heavy atoms, encompassing over 970 million unique compounds. The subset was downselected by focusing on molecules relevant to fuel applications, tailoring the dataset to the specific objective of fuel design and lowering computational overhead. Specifically, compounds composed solely of carbon (C), hydrogen (H), and oxygen (O) atoms were selected, and the molecular size was limited to those containing up to 10 heavy atoms. This initial refinement yielded a dataset of 418,274 unique species, providing a diverse representation of potential fuel molecules. This dataset was further pruned by removing species containing more than two cyclic rings. Highly polycyclic compounds introduce significant structural complexity and are less pertinent to practical fuel formulations. This additional filtering ensures that the dataset remains focused on molecules that are both computationally tractable and relevant to fuel chemistry.

RON data was further integrated to enhance the overall dataset's applicability to fuel property prediction tasks. In particular, a comprehensive RON dataset was curated from the latest literature tabulations in~\cite{Ref7,Ref18,Ref19}, which report experimentally measured single-compound RON values drawn from literature sources that used standard research-octane test procedures. The most recent reliable value of RON was retained where sources differed. The curated dataset spans RON values from below zero to above 100; sub-zero entries correspond to long chain alkane low-octane species and were retained as reported. This data curation, applying the same down-sampling criteria as those used for the GDB-13 dataset, resulted in a dataset comprising 332 species along with their RON values. It is noted that some of the species in the RON dataset were not present in the downselected GDB-13 subset. These missing species were incorporated into the dataset to ensure comprehensive coverage and maximize the model's ability to learn structure-property relationships. With this incorporation, the final dataset contained 357,907 species.

Molecular structures were represented using the SMILES notation, which encodes complex molecular graphs into linear strings suitable for computational processing. These SMILES strings were converted into one-hot encoded representations to serve as inputs to the Co-VAE model. The one-hot encoding process involved mapping each character in the SMILES string to a unique position in a binary vector, resulting in a fixed-length, two-dimensional array. All the SMILES strings were padded to a maximum length of 23 characters to maintain consistency, and the character set comprised eight unique symbols representing atoms and molecular substructures~\cite{Ref13}. This encoding ensures that the input data is uniform in size and compatible with the Long Short-Term Memory (LSTM) networks employed in the Co-VAE architecture (discussed in the next subsection). One-hot encoding for SMILES strings facilitates direct reconstruction of molecular structures from the latent space embeddings generated by the Co-VAE model. This approach offers advantages over traditional molecular descriptor-based methods by enabling an interpretable transformation between molecular structures and their encoded representations without requiring similarity search algorithms.

\subsection{Model Architecture}
The Co-VAE model developed here builds upon the authors' previous work, designed for molecular generation and property prediction. The original VAE architecture consists of a two-layer LSTM network~\cite{Ref20}, serving as the encoder, which is chosen for its ability to capture sequential dependencies in one-hot encoded SMILES strings that represent molecular structures. The encoder processes the input sequences and funnels the information through two fully connected (FC) layers into the latent space, producing means and log-variances that define the distribution from which the latent variables are sampled. The decoder mirrors this structure in reverse, mapping the latent variables back to the original dimensionality through fully connected (FC) layers and reconstructing the input sequences using an LSTM network over the sequence length. For a more detailed description of the model architecture, the readers are referred to Ref. ~\cite{Ref13}.

\begin{figure}[ht]
    \centering
    \includegraphics[width=0.5\textwidth]{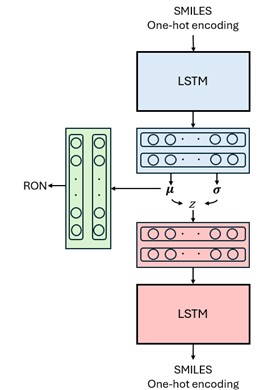} 
    \caption{A schematic of the Co-VAE framework: an LSTM encoder (shown in blue) translates SMILES one-hot encoded representation into a latent space, from which an LSTM decoder (shown in pink) reconstructs the molecular one-hot encoding, while a feedforward network (shown in green) co-optimizes the latent space for RON prediction.}
    \label{fig2}
\end{figure}

In the current study, the above architecture was extended by coupling a fuel property prediction model with the generative model so that the latent representation is trained with an auxiliary RON objective, to capture latent features meaningful for accurately predicting RON. Specifically, an additional feedforward neural network is introduced with two layers that uses the means of the latent features to predict RON. This modification allows the model to learn informative latent representations for molecular reconstruction and property estimation. A schematic of the Co-VAE model is shown in Figure~\ref{fig2}. The Co-VAE training loss function includes the RON prediction loss in addition to the original reconstruction loss quantified by Binary Cross-Entropy (BCE) and the Kullback-Leibler Divergence (KLD) regularization term. The overall loss function becomes:
\begin{equation}
\text{Loss} = \text{BCE} + \beta \,\text{*KLD} + L_{\mathrm{RON}}
\end{equation}
where $\beta$ is the weight balancing the contribution of the KLD term, and $L_{\mathrm{RON}}$ represents the mean absolute error (MAE) associated with RON prediction. 

For training the Co-VAE, same optimal $\beta$-annealing schedule as identified in the authors' previous work ~\cite{Ref13} is employed, gradually increasing $\beta$ from zero to 0.25 over 75 epochs and maintaining it at 0.25 thereafter. This approach balances reconstruction fidelity and latent space regularization, thereby preventing issues such as posterior collapse, where the decoder ignores the latent variables.

\subsection{Hyperparameter Optimization}
\label{subsec:hpo}
Optimizing the hyperparameters of the Co-VAE model is crucial to enhance its performance for both molecular generation and fuel property prediction tasks. This study systematically explored a variety of hyperparameters that influence the model capacity, complexity of the latent space, and efficiency of the training process. The hyperparameters included: number of layers in the LSTM encoder and decoder, size of the LSTM layers, sizes of the fully connected and conditioning layers, dimensionality of the latent space, and batch size. The specific ranges within which these hyperparameters were varied are provided in Table~\ref{tab1}. The learning rate was kept fixed at $1\times10^{-3}$ and the number of training epochs was set to 300. Preliminary experiments indicated that the model consistently converged within this epoch range, justifying the fixed number of epochs and learning rate. The architecture's capacity - including the depth and width of the LSTM and fully connected layers - and the latent space dimensionality were optimized to fine-tune the model, enabling it to capture the relevant features necessary for accurate molecular reconstruction as well as RON prediction.

\begin{table}[ht]
    \centering
    \caption{Hyperparameter ranges tested for the Co-VAE model.}
    \label{tab1}
    \begin{tabular}{|l|c|}
        \hline
        \textbf{Parameter} & \textbf{Value} \\
        \hline
        Number of layers in LSTM & 2 -- 3 \\
        Hidden size of each LSTM layer & 64 -- 256 \\
        Fully Connected layer size & 50 -- 150 \\
        Conditioning layer size & 10 -- 100 \\
        Latent Space dimensionality & 32 -- 128 \\
        Batch size & 64 -- 256 \\
        Learning rate & $10^{-3}$ \\
        Number of epochs & 300 \\
        \hline
    \end{tabular}
\end{table}

The down-selected GDB-13 dataset was split into 95$\%$ for training, 2.5$\%$ for validation, and 2.5$\%$ for testing. For the 332-species RON dataset, 10 species were selected for the validation set and 10 for the test set. The BCE and KLD terms in the loss function are computed for the entire GDB-derived dataset, regardless of whether or not RON measurements are available. However, only molecules with experimentally measured RON values contribute to $L_{\mathrm{RON}}$. Specifically, when a molecule in a training batch has a known RON value, its RON prediction error is included alongside the BCE and KLD loss terms; for other molecules, only the BCE and KLD loss terms are non-zero. To efficiently navigate the hyperparameter space, Bayesian optimization was performed using the bayes$\_$opt package~\cite{Ref21}. This methodology is well-suited for optimizing complex functions where evaluation is costly, as it intelligently selects hyperparameter configurations based on previous results to improve performance iteratively. The optimization process was initiated with evaluations for 16 random hyperparameter sets. Following this initial exploration, sequential iterative optimization was carried out, selecting 8 points per batch and performing 40 runs. This enabled progressive refinement of the hyperparameters based on the observed model performance. The optimization metric was defined as a composite score based on the validation set, combining reconstruction accuracy, validity of generated molecules, and loss associated with RON prediction. Reconstruction accuracy measures the model's ability to recreate input SMILES strings from the latent representations accurately. Validity assesses the fraction of molecules generated from the sampled normal distribution of the latent space that is chemically plausible and adheres to established chemical valency rules, ensuring that the outputs are meaningful within the domain of chemistry. Including RON prediction loss in the metric encourages the latent space to be informative for property prediction, aligning the generative model with the target predictive task.

\begin{figure}[ht]
    \centering
    \includegraphics[width=1\textwidth]{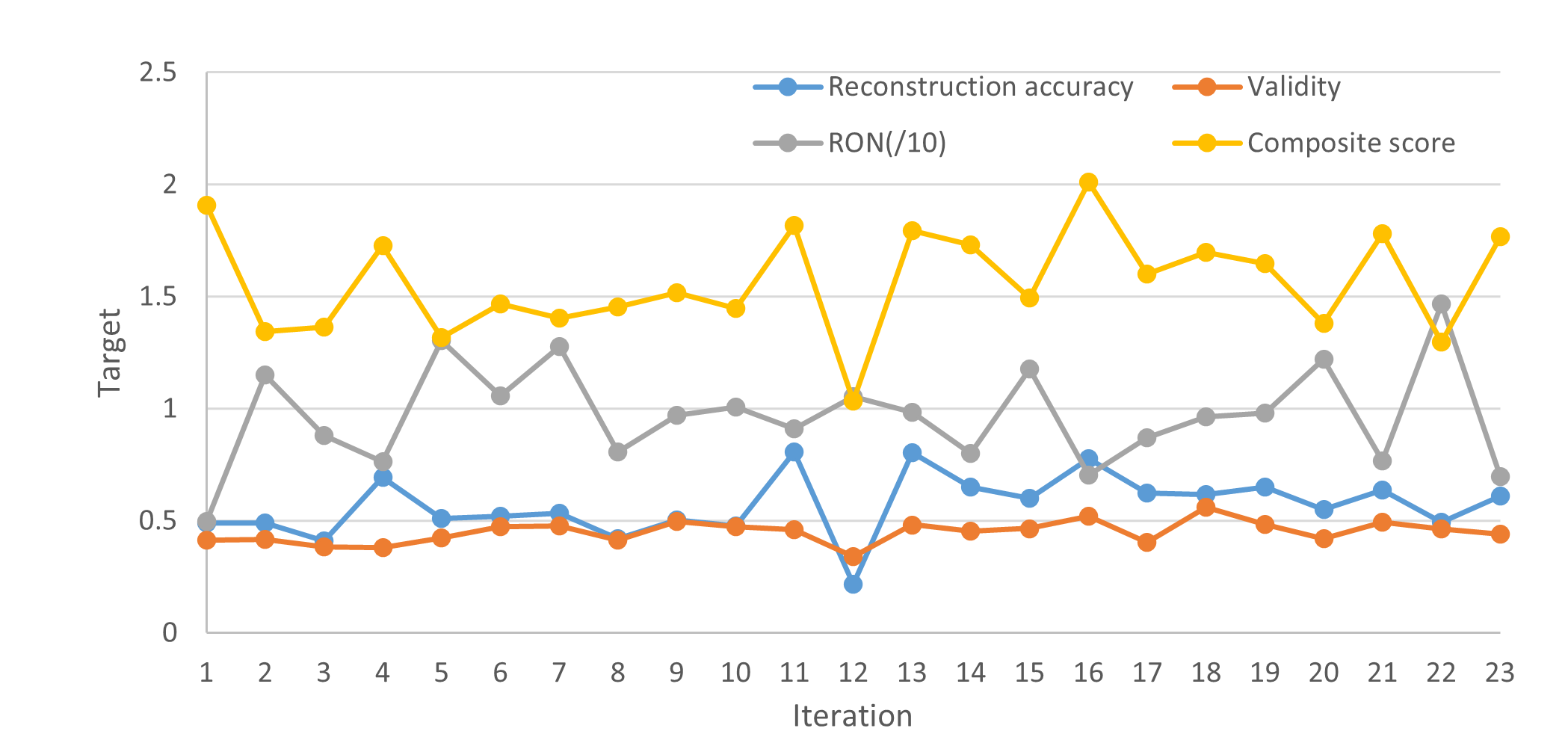} 
    \caption{Reconstruction accuracy, validity, and RON MAE (computed on validation set) for hyperparameter optimization iterations of the Co-VAE. Composite score represents the sum of reconstruction accuracy, validity, and five times the inverse of RON MAE.}
    \label{fig3}
\end{figure}

The results of the HPO are shown in Figure~\ref{fig3}, which illustrates the iterative progression of the optimization metric and demonstrates convergence toward the optimal configuration. The composite score is computed as the sum of reconstruction accuracy, validity, and five times the inverse of RON MAE to ensure comparable magnitudes. Upon completion of the optimization runs, the hyperparameter configuration that yielded the best composite score on the validation set was selected as the optimal model, corresponding to iteration 16 in Figure~\ref{fig3}. The hyperparameter values for this optimal configuration and the corresponding results on the test set are summarized in Table~\ref{tab2}.

\begin{table}[ht]
    \centering
    \caption{Best hyperparameter set obtained from HPO of the Co-VAE model and its performance on the test set.}
    \label{tab2}
    \begin{tabular}{|l|l|}
        \hline
        \textbf{Parameter} & \textbf{Value} \\
        \hline
        Number of layers in LSTM & 2 \\
        Hidden size of LSTM layer & 151 \\
        Fully Connected Layer 1 size & 84 \\
        Fully Connected Layer 2 size & 72 \\
        Conditioning Layer 1 size & 49 \\
        Conditioning Layer 2 size & 19 \\
        Latent Space size & 73 \\
        Batch size & 172 \\
        Reconstruction accuracy & 77.56\% \\
        Validity & 55.19\% \\
        RON MAE & 9.26 \\
        \hline
    \end{tabular}
\end{table}

\section{Regression Model Development for RON Prediction}
\label{sec:ronregression}
To further enhance the accuracy of RON prediction, a separate regression model was developed using the latent space embeddings from the optimized Co-VAE model. Leveraging latent space embeddings from the Co-VAE model, a variety of regression algorithms were explored to determine the most effective model for RON prediction. This approach enables us to leverage the rich, informative features captured in the Co-VAE latent space while tailoring the regression model to best fit the specific characteristics of the RON prediction task. In addition, it allows rigorous HPO to be performed beyond the constraints of the feedforward neural network within the Co-VAE architecture. The RON dataset used for the regression task was the same as the one described earlier in Section 2.1. The training set was subjected to 10-fold cross-validation to assess model performance across multiple data splits, thereby mitigating overfitting. The remaining data from the RON dataset was employed as the test set for evaluating predictive accuracy of the regression model. 

\begin{table}[htbp]
    \centering
    \caption{Regression models and hyperparameter ranges explored for RON prediction using latent space embeddings from the Co-VAE model.}
    \label{tab3}
    \begin{tabularx}{\textwidth}{lX}
        \toprule
        \textbf{Model} & \textbf{Hyperparameters} \\
        \midrule
        XGBoost & 
        \texttt{n\_estimators} [50, 1000], \texttt{max\_depth} [1, 15], \texttt{learning\_rate} [0.001, 0.3], \texttt{gamma} [0, 5], \texttt{min\_child\_weight} [1, 10], \texttt{subsample} [0.5, 1], \texttt{colsample\_bytree} [0.5, 1], \texttt{reg\_alpha} [0, 1], \texttt{reg\_lambda} [0, 1] \\
        CatBoost & 
        \texttt{iterations} [50, 1000], \texttt{learning\_rate} [0.001, 0.3], \texttt{depth} [1, 10], \texttt{l2\_leaf\_reg} [1, 10], \texttt{rsm} [0.5, 1], \texttt{bagging\_temperature} [0, 1], \texttt{border\_count} [32, 255] \\
        LightGBM & 
        \texttt{n\_estimators} [50, 1000], \texttt{learning\_rate} [0.001, 0.3], \texttt{num\_leaves} [31, 256], \texttt{max\_depth} [-1, 15], \texttt{min\_child\_samples} [5, 100], \texttt{subsample} [0.5, 1], \texttt{colsample\_bytree} [0.5, 1], \texttt{reg\_alpha} [0, 1], \texttt{reg\_lambda} [0, 1], \texttt{min\_split\_gain} [0, 1] \\
        RandomForest & 
        \texttt{n\_estimators} [50, 500], \texttt{max\_depth} [3, 20], \texttt{min\_samples\_split} [2, 10], \texttt{min\_samples\_leaf} [1, 10], \texttt{max\_features} [0.1, 1.0] \\
        GradientBoosting & 
        \texttt{n\_estimators} [50, 500], \texttt{learning\_rate} [0.01, 0.3], \texttt{max\_depth} [3, 10], \texttt{min\_samples\_split} [2, 10], \texttt{min\_samples\_leaf} [1, 10], \texttt{subsample} [0.5, 1.0], \texttt{max\_features} [0.5, 1.0] \\
        LinearRegression & 
        \texttt{fit\_intercept} [0, 1], \texttt{copy\_X} [0, 1] \\
        Ridge & 
        \texttt{alpha} [0.1, 100], \texttt{fit\_intercept} [0, 1], \texttt{solver} \{auto, svd, cholesky, lsqr, sparse\_cg, sag, saga\}, \texttt{max\_iter} [100, 1000] \\
        Lasso & 
        \texttt{alpha} [0.0001, 1], \texttt{fit\_intercept} [0, 1], \texttt{max\_iter} [100, 1000], \texttt{selection} \{cyclic, random\} \\
        ElasticNet & 
        \texttt{alpha} [0.0001, 1], \texttt{l1\_ratio} [0, 1], \texttt{fit\_intercept} [0, 1], \texttt{max\_iter} [100, 1000], \texttt{selection} \{cyclic, random\} \\
        SVR & 
        \texttt{C} [0.1, 100], \texttt{epsilon} [0.001, 1], \texttt{gamma} [0.0001, 1], \texttt{kernel} \{linear, poly, rbf, sigmoid\}, \texttt{degree} [2, 5] \\
        KNeighbors & 
        \texttt{n\_neighbors} [1, 50], \texttt{weights} \{uniform, distance\}, \texttt{algorithm} \{auto, ball\_tree, kd\_tree, brute\}, \texttt{leaf\_size} [10, 50], \texttt{p} [1, 2] \\
        DecisionTree & 
        \texttt{max\_depth} [3, 20], \texttt{min\_samples\_split} [2, 20], \texttt{min\_samples\_leaf} [1, 20], \texttt{max\_features} [0.1, 1.0], \texttt{splitter} \{best, random\}, \texttt{ccp\_alpha} [0.0, 0.1] \\
        TabNet & 
        \texttt{n\_d} [8, 64], \texttt{n\_a} [8, 64], \texttt{n\_steps} [3, 10], \texttt{gamma} [1.0, 2.0], \texttt{n\_independent} [1, 5], \texttt{n\_shared} [1, 5], \texttt{momentum} [0.01, 0.4], \texttt{lambda\_sparse} [1e-4, 1e-2], \texttt{lr} [1e-3, 1e-1] \\
        \bottomrule
    \end{tabularx}
\end{table}

In this work, we investigated a range of model classes—including ensemble-based methods, linear regression techniques, regularization, deep learning models (TabNet)~\cite{Ref22, Ref23} as well as SuperLearner approaches~\cite{Ref11, Ref24, Ref25, Ref26}—each of which offers unique advantages for capturing diverse patterns and relationships in the data. Comprehensive hyperparameter optimization was performed for each candidate regression model using NSGA-2~\cite{Ref27} from the Pymoo package~\cite{Ref28}. This multi-objective evolutionary algorithm simultaneously optimized Coefficient of Determination (R²), Root mean squared error (RMSE) and MAE, ensuring a balanced trade-off between predictive accuracy and generalization. A population size of 100 and 20 iterations were employed during the NSGA-2 optimization, which was carried out over the complete hyperparameter search space (see Table~\ref{tab3}) with 10-fold cross-validation to mitigate overfitting. For each model class, the candidate with minimum MAE value from the Pareto front was selected as the best candidate. The candidate selection is mathematically formulated as follows:
\begin{equation}
x^* = x_{\arg\min_i F_i^{(1)}}
\end{equation}
Here, \( F_i^{(1)} \) represents the MAE for the \( i^\text{th} \) candidate, and \( x^* \) denotes the optimal hyperparameter configuration corresponding to the minimum MAE value. Among all the evaluated models on the RON test set, the CatBoost model emerged as the top performer for RON prediction based on latent space representations, achieving an MAE of 5.365. 

\begin{figure}[ht]
    \centering
    \includegraphics[width=0.8\textwidth]{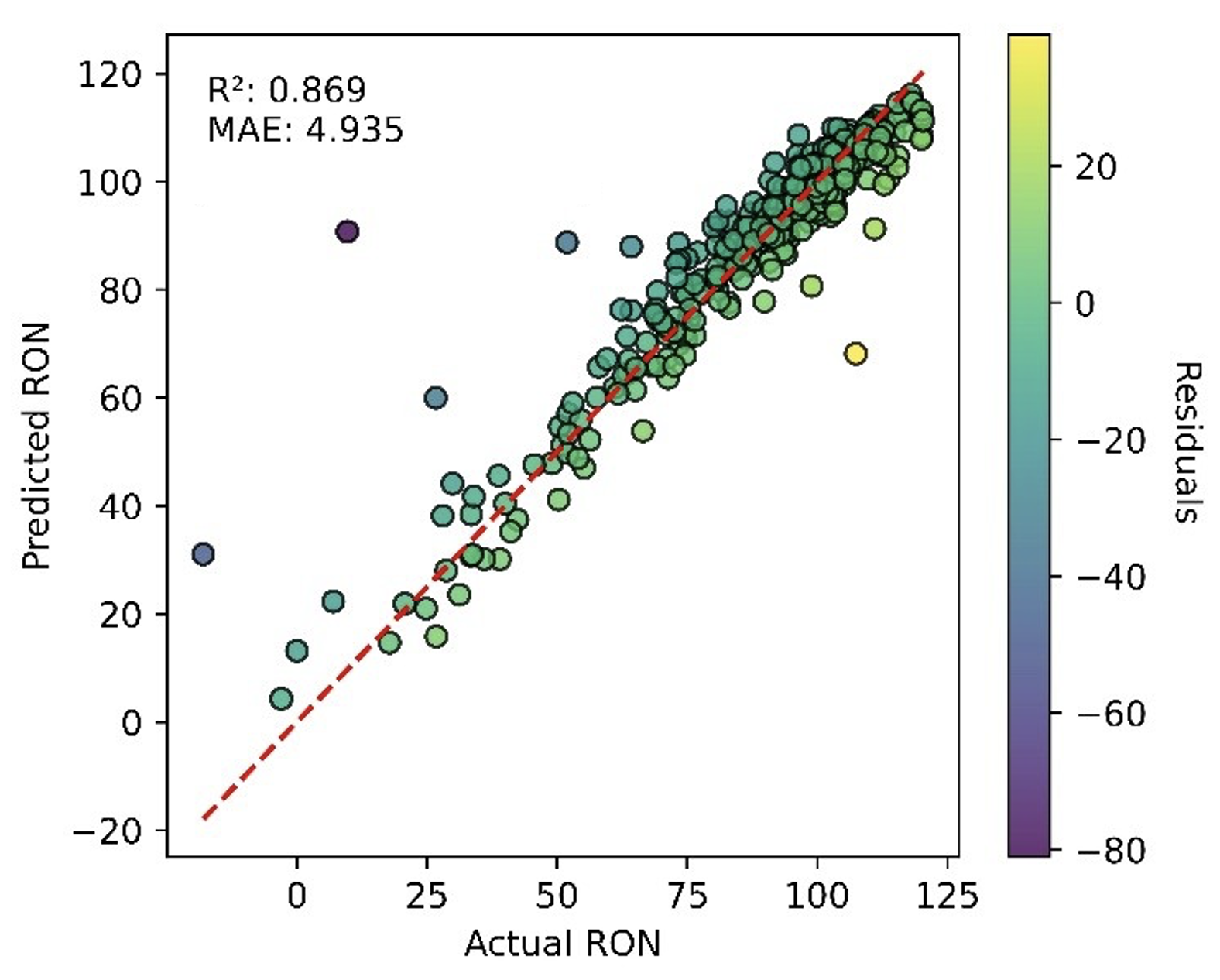} 
    \caption{Parity plot for the final optimized CatBoost model. The R²/MAE shown is the mean from the 10-fold cross-validation study.} 
    \label{fig5}
\end{figure}

To further ensure the robustness of the final model, an additional 10-fold cross-validation was performed on the entire RON dataset using the CatBoost model, configured with the best hyperparameters obtained from the multi-objective optimization. The final optimized regression model demonstrated a cross-validation performance of MAE = 4.935 as shown in the parity plot in Figure~\ref{fig5}. This level of accuracy is on par with state-of-the-art ML-based QSPR models in the literature ~\cite{Ref7, Ref19} for RON prediction. However, as discussed in the next section, the optimized RON regression model can be deployed along with the trained VAE model and a global search strategy for generative design and screening of novel high-RON fuel molecules, which is a significant advantage over traditional data-driven QSPR models. It is also noted that although the use case demonstration in this work focuses on RON as the target fuel property, the overarching data-driven pipeline can be readily extended to other (or multiple) fuel properties.

\section{Generative Design of High-RON Fuel Molecules}
\label{sec:generation}
 Based on the optimized Co-VAE and regression model developed in the previous Sections 2 and 3, the latent space of the Co-VAE was searched to identify novel fuel molecule candidates with a high Research Octane Number (RON). Specifically, the objective of the screening process was to generate molecular candidates with predicted RON values that exceeded a chosen threshold value of 110, thereby exploring the feasibility of advanced high-octane fuels. To this end, it was necessary to define the appropriate ranges for each latent dimension to systematically explore this space. The minimum and maximum values for each latent variable were computed across the entire RON dataset, which provided the natural limits of the latent space as learned by the model. To facilitate the exploration of regions just beyond the observed data, which may contain novel and promising molecules, each latent variable's range was extended by 10$\%$ of its respective span. This expansion of the search space ensures that the optimization algorithm is not constrained strictly within the confines of the existing data distribution but is still guided by the learned latent structure of chemically relevant molecules.

\begin{figure}[htbp]
    \centering
    \includegraphics[width=0.9\textwidth]{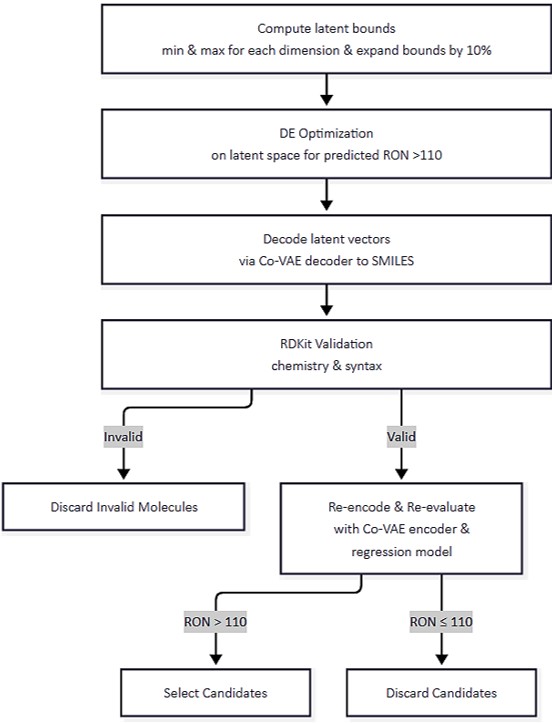} 
    \caption{Workflow for generative fuel design using Co-VAE and regression models.}
    \label{fig6}
\end{figure}

To identify latent vectors corresponding to molecules with predicted RON higher than 110, the Differential Evolution (DE) algorithm~\cite{Storn1997}—a population-based, stochastic optimization method designed for continuous and high-dimensional spaces —was employed. DE utilizes mutation, crossover, and selection processes, inspired by natural evolution, to maximize the predicted RON from the regression model, subject to defined bounds of the latent space. The SciPy implementation of DE was leveraged to perform this optimization~\cite{Ref29}. Once latent vectors with high RON values were identified, they were decoded into corresponding molecular structures using the decoder component of the Co-VAE model, which transforms each latent vector into a sequence of tokens representing a SMILES string. Thereafter, the generated SMILES strings underwent a two-step process (chemical validity check and internal consistency check) to ensure chemical feasibility and consistency in the predicted RON. First, RDKit, an open-source cheminformatics library~\cite{Ref30}, was used to verify that each generated SMILES string adheres to chemical valency rules and proper syntax, discarding any that fail this test. Beyond this initial screening, each decoded molecule was re-encoded using the Co-VAE encoder and re-evaluated with the regression model to confirm if its predicted RON remains above 110. This extra consistency check is essential because the VAE generates latent representations by sampling from a distribution, meaning the latent vector passed to the decoder is effectively a perturbed version of its ideal representation. Such perturbations can lead to fluctuations in the predicted RON, even if the molecule is chemically sound. This inherent variability—central to the variational approach—broadens chemical space exploration and necessitates a dual screening process to ensure the selection of robust, high-performance fuel candidates. In other words, the second step filters out decoding artifacts, so that only high-performance fuel candidates are selected for downstream experimental evaluation. The complete workflow for the aforementioned generative fuel design is depicted in Figure~\ref{fig6}.

\begin{figure}[htbp]
    \centering
    \includegraphics[width=1\textwidth]{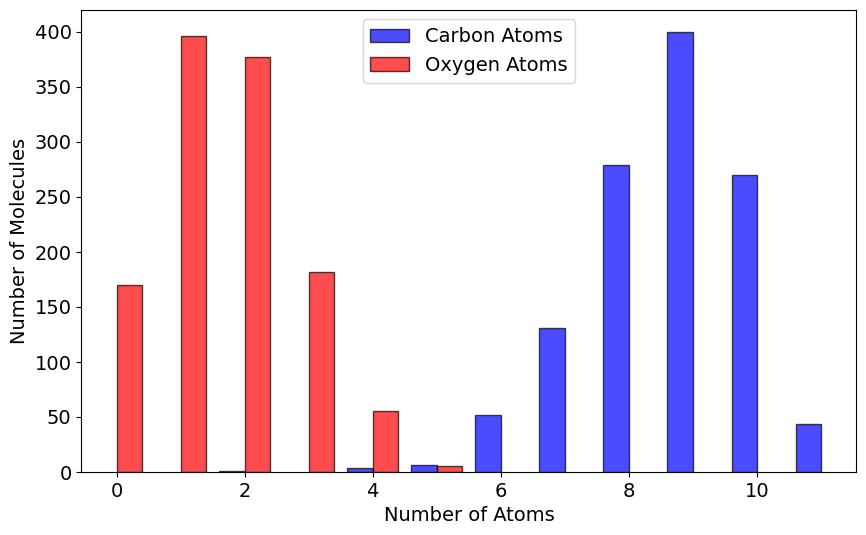} 
    \caption{Distribution of carbon and oxygen atoms in the generated molecules.}
    \label{fig7}
\end{figure}

\begin{figure}[htbp]
    \centering
    \includegraphics[width=0.8\textwidth]{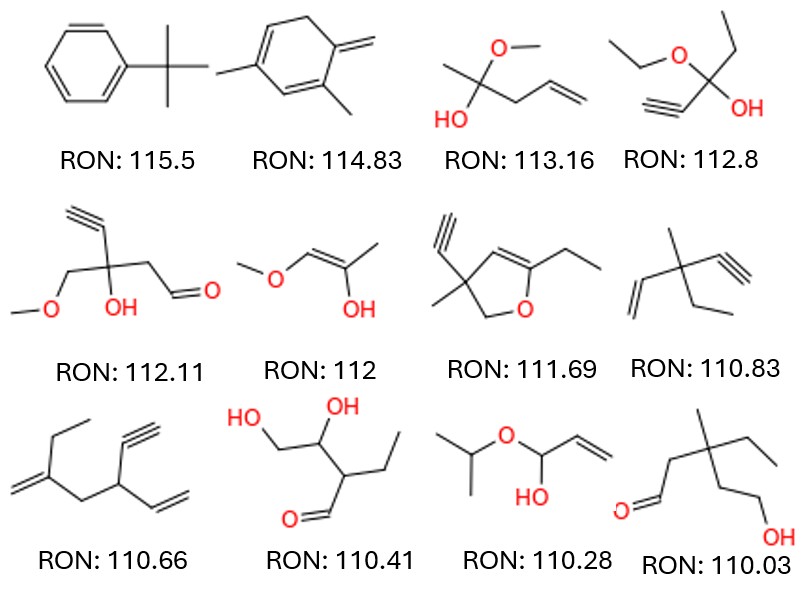} 
    \caption{A sample set of generated molecules with RON $>$ 110.}
    \label{fig8}
\end{figure}

The combination of DE search, Co-VAE decoding, and two-step screening yielded a set of 1185 unique and valid SMILES strings with estimated RON greater than 110. These represent 1185 unique chemical species. Among these species, 264 were already present in the Co-VAE training set; of these, two were included in the RON regression training dataset. Therefore, the process generated 921 unique species that the Co-VAE model had not seen. Fingerprint-based comparison against the full RON-labeled dataset (Morgan fingerprints) gives a median nearest-neighbor Tanimoto similarity of 0.242 (interquartile range: 0.206--0.303). This indicates that the generated molecules remain chemically related to known RON-relevant structures but are not direct reproductions of labeled examples. It is essential to note that these molecules were obtained from a single instance of the latent-space screening procedure. Due to stochastic nature of the DE algorithm, which is sensitive not only to initial starting conditions and prone to local convergence but also to the number of iterations and chosen population size, both of which influence the width and depth of the search, the candidate sets obtained can vary between different optimization runs. While a more exhaustive exploration of the latent space might uncover an even broader array of high‑performance fuel molecules, such a comprehensive search was out of the scope of this work and will be pursued in future work. As illustrated in Figure~\ref{fig7}, the generated molecules present a wide range of carbon and oxygen atom counts, highlighting the elemental diversity among the high‑RON candidates. Furthermore, Figure~\ref{fig8} shows a sample set of generated molecules exhibiting RON $>$ 110, which predominantly features branched structures, alcohols, ethers, and aldehydes—functional groups known to enhance RON. The generated set also exhibits high internal diversity (sampled pairwise diversity of 0.889), and 85.7\% of molecules contain oxygen, consistent with oxygenate-rich motifs often considered as high-octane candidates. This sample set not only reflects the depth of the current methodology for exploration of the latent space, but also demonstrates the effectiveness of the screening approach, revealing the diversity inherent in the search process for high‑performance fuel molecules.

It is noted that chemically valid SMILES do not imply synthesizability, stability, or safety; for a holistic downselection, additional filters for synthetic accessibility, stability, and structural alerts need to be applied, which will be pursued in future studies. Moreover, while indicative, the predicted RON values are based on the regression model trained on available data. Experimental validation is, therefore, crucial for confirming the actual octane numbers and evaluating other critical fuel properties, such as energy density, volatility, and compatibility with engine materials.

\section{Conclusions}
\label{sec:conclusions}
In this study, a Co-VAE framework was developed and applied to molecular design of fuels, with RON as the target property for a use case demonstration. By integrating a fuel property regression head with the VAE, the latent representation is trained with an auxiliary RON objective to capture features relevant to both molecular reconstruction and fuel property estimation. Through rigorous hyperparameter optimization, the Co-VAE architecture was fine-tuned to maximize molecule reconstruction accuracy, validity of generated molecules, and RON prediction capability. Leveraging latent space embeddings, a separate regression model was subsequently developed to further improve the accuracy of RON prediction. Utilizing differential evolution within the learned latent space, chemically valid novel molecules with predicted RON values exceeding 110 were then screened, illustrating the feasibility of the integrated generative and predictive modeling approach.

This framework can help identify promising fuel molecules by directing the search toward property-optimized regions of the latent space, reducing the need for random sampling from large chemical spaces. As a useful starting point aimed at RON, the demonstrated pipeline should be interpreted with its current limitations in mind. The small RON-labeled dataset relative to the molecular structural corpus limits the extent to which the latent space can be tailored to RON, and results should be interpreted as model-guided candidates rather than validated high-RON molecules— i.e., predictor bias cannot be ruled out. Predictions in high-RON regions may involve extrapolation, and the model's applicability domain is not formally characterized. Evaluation here is based on the same regression model used for optimization, so candidates are validated only in a model-internal setting; independent validation (e.g., external QSPR, physics-based estimates, or experimental evaluation) remains important as follow-on steps. In addition, realistic fuel design is inherently multi-objective, involving volatility, energy density, stability, emissions, and other constraints. Extending the screening process to additional critical fuel properties—and incorporating these with uncertainty quantification and practical filters such as synthesizability, toxicity, and stability—would enable a more comprehensive assessment of the viability of generated fuel candidates.

To enhance the proposed inverse fuel design framework, future work will focus on fine-tuning the downselection process of the GDB-13 database to curate a dataset enriched with fuel-like species, applying transfer learning from larger property-rich datasets, and extending the framework to multi-component blends where non-linear interactions become important. In addition, standard methods for measuring RON have experimental repeatability range of approximately ±1 RON; the current data-driven approach exceeds this threshold, reinforcing the need for larger and more diverse datasets to improve prediction accuracy.

\section*{Acknowledgments}
The submitted manuscript has been created by UChicago Argonne, LLC, Operator of Argonne National Laboratory (Argonne). The U.S. Government retains for itself, and others acting on its behalf, a paid-up nonexclusive, irrevocable worldwide license in the said article to reproduce, prepare derivative works, distribute copies to the public, and perform publicly and display publicly, by or on behalf of the Government. This work was supported by the U.S. Department of Energy (DOE), Office of Science under contract DE-AC02-06CH11357. GPU computing resources
from Argonne’s Laboratory Computing Resource Center (LCRC) cluster Swing and Leadership Computing Facility (ALCF) supercomputer Polaris were used for this study. The authors would like to acknowledge funding support from Aramco Americas – Detroit for this work.

\end{document}